\documentclass{article}
\usepackage{microtype}
\usepackage{graphicx}
\usepackage{subfigure}
\usepackage{booktabs} 
\usepackage{times}
\usepackage{latexsym}
\usepackage[utf8]{inputenc}
\usepackage{caption}
\usepackage[compact]{titlesec}
\usepackage[normalem]{ulem}
\usepackage[most]{tcolorbox}
\usepackage{xcolor}
\usepackage{pifont}
\usepackage{mdframed}
\usepackage{amsmath}
\usepackage{amssymb}
\usepackage{amsthm}
\usepackage{rotating} 
\usepackage{hyperref}
\usepackage{cuted} 
\usepackage[arxiv]{icml2023}
\definecolor{darkgreen}{rgb}{0.0, 0.4, 0.0}
\newcommand{\step}{\textless STEP\textgreater}
\newcommand{\lt}{\textless Line Tool\textgreater}
\newcommand{\pt}{\textless Perpendicular Tool\textgreater}
\newcommand{\pbt}{\textless Perpendicular Bisector Tool\textgreater}
\newcommand{\inrt}{\textless Intersect Tool\textgreater}
\newcommand{\ct}{\textless Circle Tool\textgreater}
\newcommand{\inlineimg}[1]{%
    \raisebox{-0.3\height}{\includegraphics[height=3em]{#1}}
}
\icmltitlerunning{Beyond Lines and Circles: Unveiling the Geometric Reasoning Gap in Large Language Models}
\DeclareCaptionLabelSeparator{emdash}{\textemdash}
\DeclareCaptionLabelFormat{customdialogue}{Generated Responses #2:} 
\newtcolorbox{dialog}[2][]{
  colback=white, 
  colframe=black, 
  boxrule=0.2pt, 
  enhanced,
  overlay={
    \node[anchor=south east, inner sep=1pt, outer sep=1pt, text=#2] at (frame.south east) {\footnotesize #1};
  }
  top=0pt,
  bottom=0pt
}
\newcommand{\cmark}{\ding{51}} 
\newcommand{\xmark}{\ding{55}} 

\begin{document}
\twocolumn[
\icmltitle{Beyond Lines and Circles: Unveiling the \\
Geometric Reasoning Gap in Large Language Models}

\icmlsetsymbol{equal}{*}
\begin{icmlauthorlist}
\icmlauthor{Spyridon Mouselinos}{uw}
\icmlauthor{Henryk Michalewski}{equal,gdm}
\icmlauthor{Mateusz Malinowski}{equal,gdm}
\end{icmlauthorlist}
\icmlaffiliation{uw}{University of Warsaw}
\icmlaffiliation{gdm}{Google DeepMind}
\icmlcorrespondingauthor{Spyridon Mouselinos}{s.mouselinos@uw.edu.pl}
\icmlkeywords{Machine Learning, Large Language Models, Geometry, Reasoning, Simulacra, Agents, ICML}
\vskip 0.3in
]
\printAffiliationsAndNotice{\icmlEqualContribution}

\begin{strip}
    \centering
    \textbf{\includegraphics[scale=0.5]{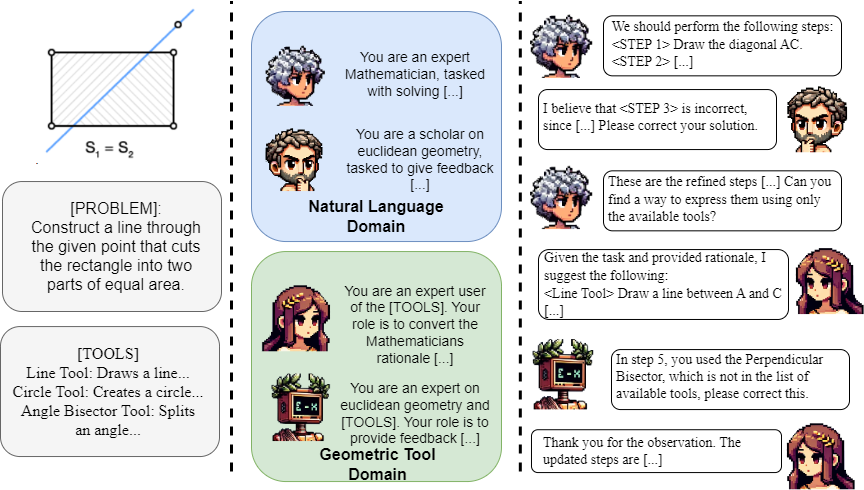}}
    \captionof{figure}{Drawing inspirations from the Ancient Greek Academy, we divide the reasoning pipeline into three stages. From left to right: The current geometric construction task is broken down into the image, its task description, and available tools. Our framework employs four LLM-based agents, each prompted with a specific role and task. A collaborative multi-round discussion is conducted where the geometric construction is effectively solved, reflecting the Academy's collective approach towards problem-solving and reasoning.}
    \label{fig:intro}
\end{strip}

\begin{abstract}
Large Language Models (LLMs) demonstrate ever-increasing abilities in mathematical and algorithmic tasks, yet their geometric reasoning skills are underexplored.
We investigate LLMs' abilities in constructive geometric problem-solving -- one of the most fundamental steps in the development of human mathematical reasoning.
Our work reveals notable challenges that the state-of-the-art  LLMs face in this domain despite many successes in similar areas. LLMs exhibit biases in target variable selection and struggle with 2D spatial relationships, often misrepresenting and hallucinating objects and their placements. To this end, we introduce a framework that formulates an LLMs-based multi-agents system that enhances their existing reasoning potential by conducting an internal dialogue.
This work underscores LLMs' current limitations in geometric reasoning and improves geometric reasoning capabilities through self-correction, collaboration, and diverse role specializations.
\end{abstract}

\section{Introduction}
Recent advancements in Large Language Models (LLMs) are groundbreaking. Models' capacity to handle complex mathematical and algorithmic tasks, interpreting data from graphs and tables, shows an ever-increasing proficiency in various forms of reasoning. Despite this, constructive geometry is one domain where LLMs still face significant challenges. This area of study, fundamental to human mathematical reasoning, lies at the intersection of tool usage, planning, and spatial reasoning.

Our investigation into LLMs' capabilities in this domain reveals several intriguing aspects. We find that in instruction following, LLMs often exhibit a bias towards the style of the examples rather than focusing on the reasoning and proper abstraction necessary for solving these geometrical problems. Furthermore, LLMs that are specialized in mathematical domains do not necessarily show proficiency in constructive geometry. This suggests that the skills required for numerical or algebraic reasoning do not directly translate to the spatial and tool-based problem-solving required in this domain. Another interesting observation is the impact of variable naming on problem-solving. Our findings show that the choice of names for variables in geometric constructions can significantly affect both the length and quality of the solutions, pointing to a potential bias in LLMs where variable names carry unintended semantic weight. Moreover,  despite being provided with visual aids, multimodal LLMs such as GPT4-V demonstrate difficulties in interpreting 2D spatial relationships. They can identify objects in a scene but struggle to integrate them into a coherent plan involving tools or steps. There's also a tendency to 'hallucinate' objects or points, which complicates their spatial reasoning process.

We propose a framework that aims to overcome these challenges. Our solution includes renaming variables to more context-appropriate names, aiming to mitigate the influence of naming conventions on the LLMs' problem-solving process. We also introduce an adaptive selection mechanism on the prompt examples, focusing the LLM on relevant information and avoiding information overload. The model uses previous instances of geometric tasks as a basis for subsequent ones, allowing it to build on past experiences and enhance its adaptability and context awareness. A key factor in our approach is the use of simulacra-based conversational agents~\cite{park2023generative}. In this setup, agents take on specialized roles, some acting as reasoners while others acting as solvers or tool users. This cross-domain dialogue leverages the strengths of each agent type and fosters a more effective problem-solving approach than traditional role-playing methods. 
Here, we aim to bridge the gap in geometric reasoning for LLMs, enhancing their capabilities in this complex and fundamental domain.\newline\textbf{Contributions.}
The main contributions of our work can be summarized in three points:\newline
\textbf{\textit{First}}, we are the first to provide an extensive analysis of the state-of-the-art leading LLMs' surprising difficulties in solving fundamental constructive geometric problems, highlighting a critical gap in their reasoning capabilities.
\newline\textbf{\textit{Second}}, we introduce three methods that assist LLMs in overcoming current limitations in the domain of constructive geometry. Our dynamic prompting mechanism enables LLMs to build upon their previous interactions instead of uninformative static prompts, our variable renaming technique neutralizes biases from variable name conventions that complicate reasoning, and our scene description prompt enhances LLMs' abilities to understand and manipulate spatial relationships in a geometric context.\newline
\textbf{\textit{Third}}, we present a novel simulacra-based system that effectively combines tool usage, instruction following, and geometric problem-solving. Our suggestions show a substantial improvement over non-collaborative methods, leading to effective solutions in geometric reasoning.
\section{Related Work}
\label{sec:related_work}
Our approach is inspired by works of various research directions, which we briefly describe here.
\newline\textbf{Prompt Engineering}
The emergent reasoning and in-context abilities of multi-billion-parameter LLMs \cite{gpt4, gpt3, bard, bloom, palm} transformed prompt engineering from simple commands to sophisticated interactions for eliciting detailed responses. Works like \cite{cot, zcot, acot} use intermediate reasoning steps in prompts, significantly improving performance in arithmetic and symbolic reasoning tasks. As an alternative to hand-crafted prompts, a series of techniques \cite{promptprogramming,largeprompt,autoprompt} are proposed for automated prompt generation, exhibiting better results in reasoning tasks. In multi-agent scenarios, \cite{camel} introduce Inception Prompting, a method that enables LLM-based agents to prompt each other in collaborative environments under the assignment of roles. Recently, \cite{Hao2023ReasoningWL,xie2023selfevaluation, tot} use tree-search and self-evaluation for tasks requiring exploration and strategic lookahead. \cite{react} unifies planning and acting in LLMs, prompting the models to generate both reasoning traces and task-specific actions. As a direct expansion, \cite{mcmc} advances these concepts with a Monte Carlo tree search over possible reasoning steps and actions, achieving state-of-the-art results in coding tasks.

\textbf{Simulacra - Conversational Agents}
The concept of 'Agents' as entities exhibiting emergent intelligence through collective interaction was introduced by \cite{som}. This idea has been extensively applied in machine learning, particularly within reinforcement learning \cite{Comback,Maac,EmergenceSym,EmergenceLang,OpenProb,Hanabi,Struct,Lmdiag,Topology}. Given their global knowledge and conversational skills, Large Language Models (LLMs) have been considered potential agents \cite{Monologue,LMAgent,CheapTalk}. \cite{socialsimulacra}, has shown that LLMs can effectively function in complex social scenarios. Recently, \cite{camel, metagpt} systematize the concept of a simulacrum, providing a less error-prone and message-effective framework for agents. Specialized agents are also suggested, with LLMs adopting different roles through prompting in various domains. Building on this work, \cite{communicative} exhibit impressive results through collaborative personas, imitating the entire workflow of a software company. Novel frameworks \cite{autogen,autoagents,agentsims, agents} add further functionalities such as visualization capabilities, the addition of custom roles and objectives, dynamic generation of agents, and distributed action execution. Finally, \cite{rolellm} introduce the first benchmark for fine-grained role-playing and suggest training on role-specific contexts and instructions, allowing models to better adapt to different characters or personas during inference.

\textbf{Geometric problems}
While mathematics and algorithms remain predominant in reasoning challenges, the exploration of geometry has been limited. Key datasets such as \cite{geos,unigeo,pgsp9k,geometry3k} typically adopt a multiple-choice format featuring annotated diagrams for interpretation and questions presented in either natural or symbolic language. Common approaches involve converting such problems into relational sets in a domain-specific language (DSL) or as formal structural clauses. The final reasoning stage is often executed through a symbolic solver or a model trained on DSL. The most recent and parallel work, AlphaGeometry \cite{AlphaGeometry}, achieved impressive results on IMO-level geometry problems. The system uses an LLM trained on synthetic DSL data as an interface to a theorem-proof engine where mathematical reasoning is delegated. Through rounds of interaction and parallel search, the LLM translates the final answer back to human-readable steps, achieving the interpretability of the proposed solution. That level of competence is possible as geometry is complete and decidable \cite{tarski1959elementary}. Contrary to that, we define all the necessary modules that generates and verify hypotheses using open-ended LLMs; showing how to improve their geometric abilities without changing their weights.
In the field of constructive geometry, which encompasses tasks requiring planning, reasoning, and the use of tools. We draw inspiration from \cite{geometry_reasoning2021, EuclidNet}, which focus on Euclidea  \cite{euclidea, euclidea_python}, a dataset that contains progressively challenging geometric problems. 
In our work, we propose an alternative to symbolic solvers and tree-based search algorithms, enhancing the reasoning capabilities of LLMs in this under-explored domain.

\section{Preliminaries}
In this section, we present the datasets, models, metrics, and definitions central to our experimental framework. \newline
\textbf{Euclidea} Our primary benchmark is the geometry game \citet{euclidea}, an online construction challenge with a range of geometric problems. The game offers eight geometric tools, the availability of which varies by level, to solve progressively complex challenges. While initial levels are straightforward, their difficulty quickly escalates, presenting significant challenges. In our experiments, we use the Python version of Euclidea \cite{euclidea_python}, designed for academic use, encompassing ninety-eight challenges across ten difficulty levels. A custom API for programmatic construction and verification of solutions is also available. We also compile a natural language version of the Euclidea dataset, including challenge definitions, solutions, and explanations from \url{https://euclidea.fandom.com/wiki/Euclidea_Wiki}, which we will make accessible for future research.

\textbf{Euclid's Elements} We explore the efficacy of training open-source LLMs on Euclid's Elements, the seminal work on constructive geometry. In Elements, a set of fundamental axioms and tools - the ruler and compass - are initially presented. Then, progressively more complex tools are synthesized through the solution of constructions using knowledge acquired exclusively from previous chapters. Similarly, various corollaries are suggested during the Elements, offering reasoning shortcuts for more advanced problems in forthcoming chapters. We posit that exposure to Elements would better align LLMs with constructive geometry tasks, as this work theoretically contains the required knowledge to solve the challenges. We utilize the publicly available English translation of Euclid's Elements \cite{elements} found \href{https://farside.ph.utexas.edu/books/Euclid/Elements.pdf}{here}.

\textbf{Models} In our experimental setup, we examine the performance of seven LLMs (five open-source and two closed-source). LLamaV2~\cite{touvron2023llama2} is one the most popular instruction-following LLMs, exhibiting impressive performance across various tasks involving commonsense reasoning, natural language understanding, dialogue, maths, and coding. We test two available checkpoints, namely its 7B and 13B variants. We also choose to include the recently introduced Mistral~\cite{mistral} and its fine-tuned variant named Zephyr~\cite{zephyr}, two 7B LLMs that, at their scale exhibit performance comparative to larger checkpoints of other open-source LLMs.
MetaMath~\cite{metamath}  specializes in mathematical and algebraic reasoning, achieving state-of-the-art results on the Math~\cite{Math} and GSM8k~\cite{Gsm8k} challenges among all open-source LLMs. We test two variants: the MetaMath-LlamaV2-13B and the MetaMath-Mistral-7B. Incorporating these models will enable us to test whether the knowledge acquired by exposure to standard mathematical datasets is transferable to the domain of constructive geometrical reasoning.
Regarding closed-source models, we select OpenAI's ChatGPT and GPT-4 models due to their superior performance in reasoning and problem-solving tasks. GPT-4, in particular, is the leading model across various datasets. Additionally, these models offer the capability to maintain a large context, which is especially beneficial for storing multiple rounds of dialogue between LLMs in our multi-agent setup. We hypothesize that fostering longer dialogues can be beneficial in domains such as constructive geometry, where various paths can yield correct solutions, and minor deviations can cause significant errors. We also aimed to examine how well these models' sophisticated instruction-following abilities perform with abstract yet precise geometric tools introduced at the prompt level. Finally, the recent integration of a vision module in GPT-4 further expands our investigation, allowing us to assess the role of visual inputs in solving geometric challenges.

\textbf{Performance metrics} 
In constructive geometry, multiple reasoning paths can lead to a correct result. In some cases, even reordering the steps of a proposed solution may be possible without harming its correctness. Thus, instead of employing the accuracy metric and requiring the model to perform an exact match of the ground truth answer, we evaluate generated constructions with the pass@k metric~\cite{spoc}. Pass@k measures the existence of a correct completion among k independent generations of the model. In~\cite{evaluating}, authors propose an updated unbiased version that we adopt throughout the rest of this work. We validate the functional correctness of the generated solutions using the Euclidea Python API.
The average of ten runs with different seeds is presented for all experiments. We choose sampling temperatures of 0.2 / 0.6 for pass@1 / pass@50 during generation.

\section{Method}
This section introduces the various components of our proposed framework, each specifically designed to address the limitations of LLMs in solving constructive geometrical problems.

\subsection{Prompt Engineering for Geometric Reasoning}
For each tested geometric challenge, we prompt our LLMs with a description outlining the available geometric tools, their expected operation, and the task requirements. We employ a few-shot learning setup to enhance our models' accuracy and reduce erroneous interpretations of tool functionalities. Specifically, we maintain a memory bank of all previously encountered problems. For each new problem, we select a subset of these problems that are most relevant to the current task. This approach ensures that our model is consistently exposed to more intricate problems and diverse tools, cultivating a more nuanced and context-aware reasoning process. We refer to this mechanism as Adaptive-Shot.

Our proposed Adaptive-Shot mechanism employs a Sentence Transformer to compare the similarity between the current level's description and available tools and those of all other levels. After this initial filtering stage, the remaining examples are presented back to the model. Subsequently, the model is tasked to identify the top five most useful examples for the current challenge, integrating them into the final few-shot prompt. 
This method not only refines the model's understanding of geometric concepts but also enhances its ability to apply this knowledge effectively to new and more complex problems. An illustrated schematic of our mechanism can be found in the Appendix, Section~\ref{sec:static_vs_adaptive}.

\subsection{From Single Models To Simulacra}
In the following stage, inspired by multi-agent collaborative environments of LLM-powered agents - simulacra \cite{camel,socialsimulacra}, we utilize role-assigning prompts and employ two sets of agents. In the first agent set, which we will refer to as solvers, the natural language solver $S_{NL}$ is tasked to generate rationales for approaching the problem in natural language, and the geometric tool solver $S_{GT}$ interprets these rationales and converts them to a series of steps using exclusively the available geometric tools. Note here that $S_{GT}$ can also be used directly without the existence of $S_{NL}$, which is the default approach of a single LLM, prompted with the available geometric tool examples of the Euclidea dataset. 
The second set of agents, which we call validators, is responsible for assessing the proposed rationales and geometric tool steps, thus introducing a new layer of roles. Like solvers, validators receive domain-specific prompts, distinguishing them as natural language or geometric tool agents. However, unlike solvers who use the adaptive-shot mechanism for their few-shot examples, validators are prompted with propositions from Euclid's Elements and a static collection of incorrect examples alongside their rectifications. Depending on their domain, these are expressed in natural language or geometric tool steps. Each validator is tasked to provide feedback to its assigned solver through rounds of dialogue. We limit the number of interactions to up to five rounds for our experiments. During each round, the validator either accepts the solver's steps or suggests changes, prompting the solver to incorporate them into their solution.

\subsection{Enhancing Spatial Awareness with the Visual Relations Prompt}
Building upon the collaborative dynamic between solvers and validators, we recognize a critical limitation in their capability to conceptualize spatial relationships in geometric problems. This limitation manifests in several ways, notably through geometrically implausible actions such as attempting to connect non-aligned points with a straight line or assuming unproven relationships between objects.

To address this, we introduce a method incorporating an auxiliary Vision-Language Large Model (VLLM) to augment scene comprehension while operating within the language domain. Our approach uses the VLLM not as the primary reasoning tool but as a scene analysis instrument. We opt for GPT4-V, given its superior performance and ease of use. The process begins by feeding the VLLM with an image-problem pair, prompting it to comprehensively describe the scene's geometric elements, their interrelations, and spatial orientations. This description, which we will refer to as the Visual Relations Prompt (VRP), is added to each agent's prompt. Through this integration, we disentangle the process of spatial recognition from the challenging geometric problem-solving that the agents must perform concurrently. The proposed VRP  is also cost-efficient, as the extraction is required only once per problem and can be shared across all agents concurrently, significantly reducing the need for recurrent interactions with visual extractors. Furthermore, it is also quite flexible, enabling models that lack innate visual capabilities to utilize the VRP, enhancing their decision-making abilities. For a comprehensive visual representation of how the VRP functions within our framework, please refer to the Appendix, Section~\ref{sec:vrp}.

\subsection{Mitigating Naming Biases}
LLMs have been studied for adopting social biases found among humans \cite{wallace,ling}, or being negatively affected by language bias in their reasoning process \cite{birds, mouselinos}. We observe a similar bias arising from the terminology used for geometric entities. For instance, when tasked with constructing a target named 'E' in contexts with existing objects' A,' 'B,' and 'C,' models will create an intermediate object 'D' before proceeding to 'E.'  This tendency extends to choosing longer reasoning paths based on the alphabetical position of the target variable, leading to unnecessary complexity and incorrect solutions. Likewise, choosing a target variable earlier in the alphabetical sequence than the required minimum steps to solve the problem can introduce faulty rationales (e.g., choosing 'C' as the target of a five-step solution can lead to early stopping on an intermediate generated 'C' point, abruptly ending the construction).

Thus, we propose a straightforward yet effective strategy to address this issue: substituting the target variable with 'X,' a universal symbol for unknowns in mathematics. This change encourages models to seek the most direct solutions, reducing the influence of alphabetical ordering. Our method proves particularly beneficial in the validation stage of our multi-agent system, where it effectively minimizes the misjudgment of correct solutions and the consequent propagation of errors. An example of such behavior can be seen in the dialogue presented in Appendix Section \ref{sec:mad}.

\section{Experiments}
\label{sec:exp}
We present a comprehensive summary of our tested models' performance on the Euclidea dataset in Table \ref{tab:results}. 
Our results are composed of three different testing setups:

In \textbf{Few-shot}, apart from the task, the description of the available tools and their intended use, models are also prompted with five solved examples using geometric tools in their solutions.

In \textbf{Finetuned}, we first fine-tune all open-source models using Euclid's book "Elements." In this way, we aim to provide models with the foundational knowledge and principles essential for addressing the challenges within the Euclidea dataset. It is crucial to highlight that, aside from the tutorial levels, there is no direct overlap between Euclidea challenges and the problems included in "Elements." During testing, models use the same setup as in Few-shot.

\textbf{Simulacra} refers to our proposed multi-agent framework, equipped with adaptive few-shot mechanism, visual relations prompting, and variable renaming. In our few-shot experimental setup, all models exhibit underwhelming performance, which improves modestly after fine-tuning. This aligns with our expectation that familiarity with standard mathematical and reasoning scenarios does not ensure proficiency in constructive geometry tasks. We posit that the increased performance from fine-tuning with Euclid's "Elements" represents an upper limit of improvement achievable by open-source models, constrained by the dataset's size. MetaMath-Mistral 7B is the most promising among the open-source options, which we further examine with our multi-agent setup. Our results reveal a significant performance boost in larger models (ChatGPT / GPT4) and, notably, in MetaMath-Mistral 7B, under our proposed framework, surpassing the few-shot ChatGPT in performance. This finding underscores the adaptability and effectiveness of our approach across a spectrum of model sizes. 

Moreover, we compare against two prior studies on the Euclidea Dataset. The model proposed by \cite{geometry_reasoning2021} combines a Masked-RCNN detector with an iterative deep search algorithm. It used the Euclidea API to verify each proposed step, continuing until a solution is found within a pre-defined depth limit. 

\begin{table}[!hb]
\begin{center}
\resizebox{\columnwidth}{!}{
\begin{tabular}{lcccc}
Method & pass@1 & pass@50 \\
\midrule
\multicolumn{3}{c}{Few-Shot} \\
\midrule
LlamaV2 (7B) &  3.1 ($\pm$ 0.5) & 4.4 ($\pm$ 0.7)\\
LlamaV2  (13B) &  4.4 ($\pm$ 0.4) & 7.5 ($\pm$ 0.8)\\
MetaMath  (13B) &  4.7 ($\pm$ 0.4) & 8.1 ($\pm$ 0.8) \\
Mistral  (7B) &   5.1  ($\pm$ 0.6) & 8.7  ($\pm$ 1.1) \\
Zephyr-$\alpha$ (7B) &   5.3  ($\pm$ 0.5) & 8.9  ($\pm$ 1.1) \\
MetaMath-Mistral  (7B) &  8.9 ($\pm$ 0.7) & 13.4 ($\pm$ 1.2) \\
ChatGPT  &  11.7 ($\pm$ 1.1) & 18.6 ($\pm$ 1.5)\\
GPT4     &  21.2 ($\pm$ 1.3) & 38.3 ($\pm$ 1.4)\\

\midrule
\multicolumn{3}{c}{Finetuned} \\
\midrule
LlamaV2 (7B) &  3.7 ($\pm$ 0.6) & 5.1 ($\pm$ 0.7)\\
LlamaV2 (13B) &  4.9 ($\pm$ 0.4) & 8.7 ($\pm$ 0.8)\\
MetaMath  (13B) &  5.3 ($\pm$ 0.5) & 9.2  ($\pm$ 1.0) \\
Mistral (7B) &   6.9  ($\pm$ 0.7) & 9.7  ($\pm$ 1.1) \\
Zephyr-$\alpha$(7B) &  6.6  ($\pm$ 0.6) & 9.5 ($\pm$ 1.2) \\
MetaMath-Mistral(7B) &  9.4 ($\pm$ 0.9) & 16.2 ($\pm$ 1.3) \\
\midrule
\multicolumn{3}{c}{Ours} \\
\midrule
MetaMath-Mistral (7B) &  14.9 ($\pm$ 1.3) & 21.1 ($\pm$ 1.7) \\
ChatGPT   &  32.3 ($\pm$ 1.3) & 61.4 ($\pm$ 2.1)\\
GPT4      &  38.9 ($\pm$ 1.1)  & 67.7 ($\pm$ 2.2)\\
\midrule
FB + MetaMath-Mistral (7B) &  15.1 ($\pm$ 1.5) & 21.4 ($\pm$ 1.6) \\
FB + ChatGPT  &  35.6 ($\pm$ 1.7) & 63.5 ($\pm$ 2.2)\\
FB + GPT4     &  41.2 ($\pm$ 1.6)  & 71.2 ($\pm$ 2.0)\\
\midrule
\multicolumn{3}{c}{Supervised Visual Component + Exhaustive Search} \\
\midrule
Method & & Accuracy &\\
\midrule
\cite{geometry_reasoning2021} (LOO-levels)    &  &44.1 & \\
\cite{geometry_reasoning2021} (LOO-packs)   &  &45.5 & \\
\cite{EuclidNet} Euclid-Net       &  &75.5 & \\
\bottomrule
\end{tabular}
}
\caption{\small{Results on Euclidea dataset. Ours refers to the $S_{NL}-S_{GT}$ setup employing VRP, variable renaming and Adaptive Shot (Self). LOO refers to the "Leave-One-Out" method: For each given level in the Euclidea challenge, the model is either trained on the rest levels in the same pack (LOO-level) or the rest of packs in the dataset (LOO-packs).}}\label{tab:results}
\end{center}
\end{table}

Euclid-Net \cite{EuclidNet} employs a similar method named "Deep visual reasoning with backtracking," where a neural network assists in step selection, exploring solutions in a tree-like structure. A notable difference between these models and ours is their dependence on a precise visual component extractor and real-time, ground truth feedback during their trial-and-error process. To assess the possible benefits of real-time feedback, we introduce Feedback mode (+FB). Validators now have access to ground truth answers and are restricted to merely informing solvers about the correctness of their suggested steps. In this way, there is no solution leak to the solvers. Our findings reveal only a slight performance improvement, indicating that solvers struggle to develop more complex solutions or alternate reasoning paths even with ground truth hints. This underscores the nuanced complexities of applying LLMs to constructive geometry.

\section{Ablation Studies}
\subsection{Overcoming hallucinations and context overdependence}
In our exploration of LLMs' proficiency in geometric problem-solving, we initially used models in a zero-shot setup, with the only context being the tool definitions. This yielded subpar results, with models hallucinating the functionalities of the tools. Consequently, we shifted to a few-shot approach, integrating examples of solved problems into our prompts. This noticeably reduced hallucinations, but models seemed to over-rely on the provided examples, often replicating sequences of steps in their solutions.

We hypothesized that this behavior stems from the instruction-tuning training process: Many LLMs are fine-tuned to follow instructions closely, primarily replicating styles and patterns. In our case, models were not merely mimicking styles but inadvertently replicating entire reasoning processes. Instruction-following proved counterproductive here, especially since many geometric problems share common preliminary steps (e.g., constructing bisectors or circles of equal radius). Models frequently repeated specific groups of steps, irrespective of their relevance to the problem, leading to redundant and incorrect constructions. This is further illustrated in Table \ref{tab:ablation_few_shot}, where we evaluated the model's performance in solving geometric problems from three distinct difficulty levels: Tutorial, Alpha, and Beta. Interestingly, unlike the simpler Alpha and Tutorial levels, the model exhibited improved performance when prompted with higher difficulty (Beta) examples. This observation contradicts typical few-shot prompts in algorithmic or mathematical problem-solving, which include straightforward, demonstrative examples mirroring the dataset's style and expected logic.

Recognizing these limitations, we repeated our experiments using our proposed adaptive few-shot approach. Apart from Adaptive-Shot (Self), we also tested with a simpler variant, Adaptive-Shot (ST), where the Sentence Transformer, rather than the model, determined the selection of few-shot examples after the filtering stage. Our findings demonstrate that adaptive variations outperform the static approach, corroborating our hypothesis. The Adaptive-Shot (Self) method proved slightly more effective, albeit with the trade-off of necessitating more API calls than the Adaptive-Shot (ST), particularly in closed-source models. The method is also limited by the capacity of the model to identify what prompts would be more useful for the task, which we hypothesized to surpass the contextual abilities of an auxiliary sentence transformer.

\begin{table}[!h]
\begin{center}
\resizebox{0.75\columnwidth}{!}{
\begin{tabular}{lcc}
\cmidrule(lr){1-3}
Method &  pass@1 & pass@50 \\
\midrule
Zero-Shot              &  5.9 ($\pm$ 1.9) & 9.6 ($\pm$ 2.7)   \\
Few-Shot (Tutorial)    &  7.2 ($\pm$ 1.4) & 15.9 ($\pm$ 3.1)   \\
Few-Shot (Alpha)       &   11.4  ($\pm$ 1.8) & 18.6 ($\pm$ 3.2) \\
Few-Shot (Beta)       &   12.7 ($\pm$ 2.2) & 20.8 ($\pm$ 3.6)   \\
Adaptive-Shot (ST)     &   13.3 ($\pm$ 1.8) & 21.2  ($\pm$ 3.0)  \\
Adaptive-Shot (Self)   &   \textbf{13.6} ($\pm$ 1.8) & \textbf{21.5} ($\pm$ 2.7)  \\
\bottomrule
\end{tabular}
}
\caption{Effectiveness of in-context example methods. Experiments with ChatGPT 3.5-Turbo on the Alpha and Beta levels. }\label{tab:ablation_few_shot}
\end{center}
\end{table}

\subsection{Effectiveness of domain and role division}
Interestingly, LLMs proved quite successful when prompted to merely generate ideas - and not specific steps using geometric tools. Although they made convenient assumptions, their overall reasoning was still accurate. This led us to question the root cause of the discrepancy between knowing the solution in the form of a plan expressed in natural language and failing to execute it accurately using strict and abstract geometric tools. 

We decided to test whether we can instead benefit from the synergy of those worlds: an LLM instance could focus on generating ideas in the natural language domain, and another could specialize in transforming these ideas into a series of geometric tool steps. By employing such a multi-agent setup, as seen in Table \ref{tab:ablation_multi_agent}, we observed a significant performance gain – $13.6 \rightarrow 21.5$ pass@1 – when comparing a single agent ($S_{GT}$) operating directly with geometric tools against a duo of collaborative agents with differentiated domains ($S_{NL}-S_{GT}$). However, we noticed that the chaining of agents could sometimes result in the propagation of errors. For instance, a natural language solver producing a flawed rationale would inevitably pass these inaccuracies on to the geometric tool solver. To counter this, we experimented with the introduction of a set of validators. These agents, prompted under a different role and context, helped to minimize errors both during the transition between domains and by preventing the improper application of tools, leading to a further performance boost of $22.2 \rightarrow 28.1$ pass@1 in our proposed multi-agent configuration.

\begin{table}[!h]
\begin{center}
\resizebox{\columnwidth}{!}{
\begin{tabular}{lcccc}
\cmidrule(lr){1-5}
Configuration & Agents & Domains &  pass@1 & pass@50 \\
\midrule
$S_{GT}$            & 1 &  1   & 13.6 ($\pm$ 1.8)  & 21.5 ($\pm$ 2.7)  \\
$SV_{GT}$           & 2 &  1   & 17.9 ($\pm$ 1.2)  & 34.9 ($\pm$ 2.3) \\
$S_{NL}-S_{GT}$      & 2 &  2   & 22.2 ($\pm$ 1.3)  & 46.7 ($\pm$ 2.0) \\
$SV_{NL}-SV_{GT}$    & 4 &  2   & \textbf{28.1} ($\pm$ 1.1)  & \textbf{53.5} ($\pm$ 2.4) \\
\bottomrule
\end{tabular}
}
\caption{Effectiveness of different agent configurations: Experiments with ChatGPT 3.5-Turbo on Alpha and Beta levels. \textbf{S} refers Solver, \textbf{V} refers to Validator, \textbf{NL} to Natural Language and \textbf{GT} to geometric tool domain.}\label{tab:ablation_multi_agent}
\end{center}
\end{table}

\subsection{Visual Aids in Spatial Reasoning}
Another noteworthy finding was that LLMs would often create new geometric objects without acknowledging their overlap with existing ones. Moreover, they occasionally suggested steps that violated geometric rules or led to repetitive movements. We hypothesized that LLMs' difficulty with geometric reasoning in 2D spaces stems from a lack of exposure to such setups, typically operating in a unidimensional, left-to-right manner. This raised the question: Could introducing a visual signal bridge this reasoning gap?

To test this, we used GPT4-V, a multimodal variant of GPT4 known for its proficiency in visual question answering. We began with simple tests, prompting GPT4-V to identify freehand sketches of geometric objects. The model identified these features successfully, including subtle aspects like right angles indicated by small corner squares. We then proceeded to a more complex scenario: We presented GPT4-V with an image/problem pair and asked for the first solution step. We deliberately performed an incorrect step and found that GPT4-V often validated these errors, suggesting a disconnect between scene understanding and geometric reasoning. We present these findings in Table \ref{tab:ablation_vrp}. Comparing GPT4 with its multimodal variant, GPT4-V + Image revealed a marginal improvement, suggesting that a visual signal could only assist the initial scene understanding without significantly improving spatial arrangement in later stages. Building on these insights, we compare our proposed VRP method  (VRP-GPT4) to the multimodal approach  (GPT4-V + Image). Results demonstrated a slight performance boost with VRP, with its real advantage lying, however, in enhancing the capabilities of non-visually capable models like the tested ChatGPT and multi-agent configurations, which benefited greatly.
\begin{table}[!hb]
\begin{center}
\resizebox{0.9\columnwidth}{!}{
\begin{tabular}{c|lcc}
Domain & Method &  pass@1 & pass@50 \\
\midrule
Language& ChatGPT & 13.6 ($\pm$ 1.8)  & 21.5 ($\pm$ 2.7)  \\
&GPT4                            & 23.9  ($\pm$ 0.9)  & 44.8 ($\pm$ 1.6) \\
&ChatGPT$^\star$   & 28.1 ($\pm$ 1.2)  & 53.5 ($\pm$ 1.7) \\
&GPT4$^\star$ &  33.7 ($\pm$ 1.0)  &  62.0 ($\pm$ 1.3) \\
\midrule
Multimodal&GPT4-V + Image                     &  24.2 ($\pm$ 1.4)  & 45.1 ($\pm$ 1.7) \\
&VRP-ChatGPT         & 19.4 ($\pm$ 1.1)  &  37.1 ($\pm$ 1.3) \\
&VRP-ChatGPT$^\star$   & 34.5 ($\pm$ 0.8)  & 59.2 ($\pm$ 1.2) \\
&VRP-GPT4          & 25.3 ($\pm$ 0.9)  & 48.6 ($\pm$ 1.4) \\
&VRP-GPT4$^\star$    & \textbf{38.8} ($\pm$ 0.9)  & \textbf{64.6} ($\pm$ 1.2) \\
\bottomrule
\end{tabular}
}
\caption{Multimodal prompt effectiveness. Experiments on the Alpha and Beta pack levels. The ($\star$) symbol refers to an $SV_{NL}-SV_{GT}$ multi-agent configuration.}\label{tab:ablation_vrp}
\end{center}
\end{table}
\vspace{-16pt}

\subsection{Impact of Geometry Nomenclature on LLMs}
LLMs mirror the human convention of alphabetical naming in mathematical contexts. The choice of target variables later in the alphabetical order leads to longer and more inaccurate solutions. This is a byproduct of their training on human-generated texts, where entities in algebraic or geometric contexts typically adhere to an alphabetical naming convention -labeling a triangle as ABC rather than EOA. In the same fashion, variables associated with the assignment of solutions, like X, would theoretically condition the LLM to find its value, possibly assigning it to any given constructed object.
To empirically validate this hypothesis, we designed an experiment with 20 geometric problems, each requiring 3 to 5 solution steps. In these problems, we manipulated the target variable in three distinct ways: maintaining the original name (+0), replacing it with the letter X (+X), and renaming it using a letter 1, 2, or 3 positions further in the alphabetical sequence (+1, +2, +3). As depicted in Table \ref{tab:ablation_rename}, models perform worse when the target variable is shifted by one or two letters in the alphabet. Interestingly, this tendency diminishes when the target is more than three letters away and is further reduced with the substitution of 'X'. This underscores the effectiveness of using 'X' as a neutral, bias-mitigating variable in our multi-agent framework. 

\begin{table}[!h]
\begin{center}
\resizebox{0.9\columnwidth}{!}{
\begin{tabular}{cccc}
 &  +0 & +1 & + 2 \\
\midrule
ChatGPT          & 10.7 ($\pm$ 2.1) & 10.1 ($\pm$ 2.4) & 10.1 ($\pm$ 2.4)  \\
GPT4             & 35.2 ($\pm$ 2.5) & 32.8 ($\pm$ 2.6) & 31.6 ($\pm$ 2.8)   \\
ChatGPT$^\star$  & 48.1 ($\pm$ 2.3) & 46.3 ($\pm$ 2.1) & 47.5 ($\pm$ 1.9)    \\
GPT4$^\star$     & 52.9 ($\pm$ 2.1) & 50.4 ($\pm$ 2.2) & 50.7 ($\pm$ 2.7)   \\
VRP-ChatGPT$^\star$  & 60.7 ($\pm$ 2.9) & 57.9 ($\pm$ 2.5) & 57.9 ($\pm$ 3.0)    \\
VRP-GPT4$^\star$     & 65.1 ($\pm$ 2.2) & 63.4 ($\pm$ 3.2) & 63.6 ($\pm$ 2.9)   \\
\midrule
\multicolumn{1}{c}{} & \multicolumn{1}{c}{+3} & \multicolumn{1}{c}{+X} & \multicolumn{1}{c}{} \\
\midrule
ChatGPT         & 10.6 ($\pm$ 1.2) & 10.5 ($\pm$ 1.7) \\
GPT4            & 32.4 ($\pm$ 1.1) & 37.5 ($\pm$ 1.3)  \\
ChatGPT$^\star$  & 47.9 ($\pm$ 1.7) & 48.6 ($\pm$ 1.9)  \\
GPT4$^\star$     & 51.8 ($\pm$ 1.4) & 53.3 ($\pm$ 1.5)   \\
VRP-ChatGPT$^\star$  & 59.2 ($\pm$ 2.3) &  61.1 ($\pm$ 2.2)   \\
VRP-GPT4$^\star$     & 64.7 ($\pm$ 2.1) &  66.2 ($\pm$ 2.4)   \\
\bottomrule
\end{tabular}
}
\caption{Qualitative results on the effect of variable renaming. Results refer to the pass@50 metric. The star symbol ($\star$) refers to $SV_{NL}-SV_{GT}$ configuration.}\label{tab:ablation_rename}
\end{center}
\end{table}
\vspace{-12pt}

\subsection{Generalisation to different datasets}
To expand the scope of our research and examine the potential generalizability of our multi-agent framework, we tested its performance against three datasets involving mathematical reasoning: GSM8K, SVAMP, and the Geometry split from the MATH dataset. 
We were inspired by the findings of \cite{msel}, who identified diverse reasoning patterns when LLMs are engaged in problem-solving using Chain of Thought (COT) \cite{cot} versus Program-Aid (PAL) \cite{pal} methodologies. While COT is recognized for its creativity and flexibility in devising solutions, PAL is noted for its enhanced accuracy in numerical computations. This differentiation of roles was an ideal fit for our framework: A pair of Solver-Validator agents ($SV_{COT}$) could initially converse about solving a math problem using a chain of thoughts approach. Then, the produced rationale would be passed down to another pair of agents ($SV_{PAL}$), specialized in transforming the proposed steps into Python code solutions that can be computed precisely. We opted not to employ a Python interpreter as the second validator ($V_{PAL}$), similar to what code-based verification methods like \cite{csv} do. Instead, we introduced another LLM agent in this role, thereby maintaining consistency in our approach of utilizing LLMs across different stages of problem-solving. Finally, we used the Adaptive Few-Shot mechanism using examples from each dataset's training split.
Our results, presented in Table \ref{tab:results_extra}, show that our multi-agent setup can perform very close to current state-of-the-art methods despite not being primarily designed for these tasks. This is also true for the MetaMath-Mistral model, which also enjoys a significant performance boost using our proposed method, even in this non-geometric setup.

\begin{table}[!h]
\begin{center}
\resizebox{1\columnwidth}{!}{
\begin{tabular}{clccc}
Method &  Model & GSM8K & SVAMP & Geometry  \\
\midrule                           
 & MetaMath-Mistral (7B)  &  84.3  & 79.7  &  21.6\\
Ours                           & ChatGPT  &  88.4  & 86.1  &  40.2\\
                 & GPT4     & 96.9   & 95.8  & 56.3\\
\midrule
MetaMath  &  MetaMath-Mistral (7B) & 77.7  & 75.8  &  18.4\\
~\cite{metamath}  &   &  &  \\
MS (SC, K=15)  &   ChatGPT & 89.2  & 85.2   &  N/A \\
~\cite{msel}     &   GPT4    & 96.8  &  95.8  &  N/A \\
                             
PHP (SC, K=40)  &  ChatGPT       &  85.1  & 83.1  &  25.4 \\
~\cite{php}                  &  GPT4          &  95.5  & 91.9  &  41.9  \\
\midrule
CSV (K=1)        &  GPT4-Code      &  92.9  & N/A  &  54.0 \\
CSV (K=16)        &  GPT4-Code      &  97.0  & N/A  &  64.9 \\
  ~\cite{csv}              &  &  &  & \\
\end{tabular}
}
\caption{\small{Results of the multi-agent framework on mathematical datasets. N/A refers to not reported results.}}\label{tab:results_extra}
\end{center}
\end{table}

\section{Conclusions}
Our study highlights the intrinsic challenges that LLMs face in constructive geometry. We observe a limited skill transfer from other mathematical domains, insufficiencies of typical prompting techniques, and a lack of 2D spatial reasoning. 
We thus identify that the existing LLMs are inferior in achieving very competitive performance on geometric tasks unless we reinforce the system with strong inductive biases such as a theorem prover \cite{AlphaGeometry}. On the other hand, we also show that our multi-agent system with role specializations and interactions can substantially boost the performance of the whole system, and tackle geometric problems more effectively. That system has some properties of the conceptual integration networks \cite{fauconnier1998conceptual} where the ``blending'' happens if the system is considered as a whole.

We hope that our work provides a starting point for a longer-term view where the system is explicitly trained to interact in a multi-agent setting.
Though challenging, this paradigm shift from merely enhancing existing models to creating a new generation of LLMs will eventually be required for significant breakthroughs in domains where deep, specific, and accurate cognitive processing is crucial.

\clearpage
\section{Limitations}
In our study, we specifically focus on constructive geometry, a distinct subset within the broader field of geometry. While numerous datasets address geometric problems, they predominantly focus on Vision-Language Models (VLMs) and employ a question-answering (QA) format, which differs fundamentally from the nature of constructive geometry. To the best of our knowledge, the dataset used is unique in this area, and we hope to attract more attention to the domain. We also recognize that our multi-agent setup might not be universally applicable, particularly for general-purpose datasets and solutions. Our approach could be less effective and slower than few-shot methods in scenarios where distinct modalities or roles are unnecessary.

Furthermore, it is crucial to consider the significant costs associated with experiments involving closed-source LLMs behind APIs. The extensive communication rounds and large context sizes required by our method can lead to substantial expenses. In cases where cost is a concern, a more straightforward, single-agent approach may be more viable and cost-effective. 

\section{Risks and Ethical Considerations}
We do not identify any risks or ethical considerations associated with our proposed ideas and suggested methods.

\section*{Acknowledgements}
All experiments were performed using the Entropy cluster funded by
NVIDIA, Intel, the Polish National Science Center grant
UMO-2017/26/E/ST6/00622 and ERC Starting Grant TOTAL. The work of
Spyridon Mouselinos and Henryk Michalewski was supported by the Polish
National Science Center grant UMO-2018/29/B/ST6/02959.
\bibliography{main}
\bibliographystyle{icml2023}

\newpage
\appendix
\onecolumn
\section{Information on Models and Datasets}
\label{sec:modelsanddata}
\begin{table}[!ht]
\begin{center}
\resizebox{\columnwidth}{!}{
\begin{tabular}{lcccl}
Model Name &  Link  & LICENSE \\
\midrule
LlamaV2 \cite{touvron2023llama2} & \url{https://github.com/facebookresearch/llama} & LLAMA 2 COMMUNITY LICENSE AGREEMENT\\
MetaMath \cite{metamath}  & \url{https://github.com/meta-math/MetaMath} & Apache License 2.0    \\
Mistral \cite{mistral}  &  \url{https://github.com/mistralai/mistral-src} & Apache License 2.0\\
Zephyr-$\alpha$ \cite{zephyr}  & \url{https://huggingface.co/HuggingFaceH4/zephyr-7b-alpha} & MIT\\
MetaMath-Mistral \cite{metamath} &  \url{https://github.com/meta-math/MetaMath} &  Apache License 2.0\\
ChatGPT \cite{InstructGPT} &   OpenAI - API usage of \textit{gpt-3.5-turbo-16k} &  N/A\\
GPT4 \cite{gpt4}  &   OpenAI - API usage of \textit{gpt-4-32k} &  N/A\\
GPT4-V \cite{gpt4}  &   OpenAI - API usage of \textit{gpt-4-vision-preview} &  N/A\\
\bottomrule
\end{tabular}
}
\end{center}
\caption{URL and Licenses of used Large Language Models / APIs.}
\label{table:model_links}
\end{table}

\begin{table}[!ht]
\begin{center}
\small
\resizebox{\columnwidth}{!}{
\begin{tabular}{lcc}
Dataset Name &  Link & LICENSE  \\
\midrule
Euclid-Elements \cite{elements}& \url{https://farside.ph.utexas.edu/books/Euclid/Elements.pdf} & CC BY-NC-SA 4.0\\
Python Port of Euclidea \cite{euclidea_python} & \url{https://github.com/mirefek/py_euclidea} & MIT \\
Euclidea Wiki Page & \url{https://euclidea.fandom.com/wiki/Euclidea_Wiki} & CC BY-SA \\
GSM8k \cite{Gsm8k}& \url{https://github.com/openai/grade-school-math} & MIT \\
SVAMP \cite{svamp} & \url{https://github.com/arkilpatel/SVAMP} & MIT \\
Geometry / MATH \cite{Math} & \url{https://github.com/hendrycks/math/} & MIT \\
\bottomrule
\end{tabular}
}
\end{center}
\caption{URL and Licenses of used Datasets.}
\label{table:data_links}
\end{table}

\section{Information on Experimental Setup}
Our experimental setup consisted of 1x NVIDIA A100 GPU. Regarding the fine-tuning results on Euclid's Elements of Table \ref{tab:results}, we trained all LLMs using the bitsandbytes library (\url{https://github.com/TimDettmers/bitsandbytes}) and 4-bit quantization with the QLoRA technique \cite{qlora}. Below, the reader can find the full hyperparameter list:

\begin{table}[h]
\centering
\small
\begin{tabular}{|l|l|}
\hline
\textbf{Hyperparameter}                  & \textbf{Value}     \\ \hline
Training Epochs                       & 3                  \\ \hline
Batch Size          & 32                 \\ \hline
Accumulation Steps           & 4                  \\ \hline
Learning Rate                           & 2e-5               \\ \hline
Warmup Ratio                            & 0.03               \\ \hline
Scheduler                      & Cosine           \\ \hline
Max Gradient Norm                          & 0.3                \\ \hline
Weight Decay                            & 0.001              \\ \hline
Lora Alpha                              & 16                 \\ \hline
Lora Dropout                            & 0.1                \\ \hline
Lora R                                  & 64                 \\ \hline
Use 4bit                                & True               \\ \hline
BnB\_4bit\_compute\_dtype                & Float16          \\ \hline
BnB\_4bit\_quant\_type                   & NF4              \\ \hline
Gradient Checkpointing                  & True               \\ \hline
Optimizer                                    & Paged\_adamw\_32bit \\ \hline
\end{tabular}
\caption{Hyperparameters and their values}
\label{table:hyperparameters}
\end{table}

Regarding the training objective, we used the typical causal language modeling loss. Moreover, we used a validation split of 10\% sampled uniformly across different chapters of the book. We monitored the validation loss every 500 steps as our metric for early stopping.

Regarding the API calls to OpenAI models, gpt3.5-turbo-16k was used for the ChatGPT experiments, gpt4-32k was used for the GPT4 experiments, and for the extraction of Visual Relation Prompts, The endpoint called gpt4-vision-preview was used. All of our API calls were subject to throttling limits, and waiting loops were introduced to avoid interruptions of service. We conducted most of our ablation studies and early experiments with ChatGPT to avoid massive waiting times and reduce the high experimental cost. The total experiment time was approximately 500 hours, and our total costs were around 2000 USD.

\clearpage
\section{Tools and problems from Euclidea dataset}
There are 10 tools available in Euclidea, although not all of them can be used on every level. Tools become available progressively as the difficulty increases. 
Here is a list of all available tools:
\begin{enumerate}
\item Move Tool: Moves a geometric object.
\item Point Tool: Marks a point and labels it. 
\item Line Tool: Draws a line between two points or a ray from a starting point. 
\item Circle Tool: Constructs a circle using a specific point as the center. 
\item Perpendicular Bisector Tool: Creates the perpendicular bisector of a segment between two points. 
\item Perpendicular Tool: Draws a line perpendicular to a given one at a specific point.
\item Angle Bisector Tool: Creates a line that bisects a given angle. The line originates at the vertex of the angle. 
\item Parallel Tool: Draws a line parallel to a given line or segment.
\item Compass Tool: Uses a compass to construct a circle with a radius equal to a given segment.
\item Intersect Tool: Marks the intersection between two geometric objects.
\end{enumerate}

Here are some sample problems from different difficulty levels:
\begin{dialog}[Problem: Rhombus in Rectangle - Pack: Alpha]{gray}
Given the rectangle ABCD with $AB>AD$. Inscribe a rhombus in the rectangle so that they share a diagonal.\newline
Available Tools: [Line Tool, Circle Tool, Perpendicular Bisector Tool, Intersect Tool]\newline
Solution:\newline
Perpendicular Bisector Tool: Construct the perpendicular bisector of AC, intersecting AB at E and CD at F.\newline
Line Tool: Construct line AF.\newline
Line Tool: Construct line CE.\newline
\end{dialog}

\begin{dialog}[Problem: Lozenge - Pack: Gamma]{gray}
Given the side AB. Construct a rhombus with the given side and an angle of 45° in a vertex.\newline
Available Tools: [Line Tool, Circle Tool, Perpendicular Bisector Tool, Perpendicular Tool, Intersect Tool, Angle Bisector Tool]\newline
Solution:\newline
Perpendicular Tool: Construct the perpendicular to AB from A; let C be a distinct arbitrary point on that perpendicular.\newline
Angle Bisector Tool: Construct the angle bisector of BAC.\newline
Circle Tool: Construct the circle with center A and radius AB, intersecting the angle bisector at D and line AC at E with E on the side of A opposite from C.\newline
Perpendicular Tool: Construct the perpendicular to AC through D.\newline
Line Tool: Construct line BE.\newline
\end{dialog}

\begin{dialog}[Problem: Angle of 54 - Pack: Theta]{gray}
Let A be the vertex. Construct two rays that divide the given angle of 54 degrees into three equal parts.\newline
Available Tools: [Line Tool, Circle Tool, Perpendicular Bisector Tool, Intersect Tool, Parallel Tool, Compass Tool]\newline
Solution:\newline
Circle Tool: Construct a circle with center B on either line and radius AB. Circle B intersects with line AB at point C, and the other line at point D.\newline
Circle Tool: Construct a circle with center D and radius AD. Circle D intersects with circle B at point E.\newline
Line Tool: Draw line AE. This is the first angle trisector.\newline
Circle Tool: Draw circle with center E and radius EC. Circle E intersects with circle B at point F.\newline
Line Tool: Draw line AF. This is the second angle trisector.\newline
\end{dialog}

\clearpage
\section{Static Versus Adaptive Few-Shot}
\label{sec:static_vs_adaptive}
In this section, we will provide an analytical overview of our adaptive few-shot mechanism.   
\begin{figure}[!h]
    \centering
      \includegraphics[width=0.8\textwidth]{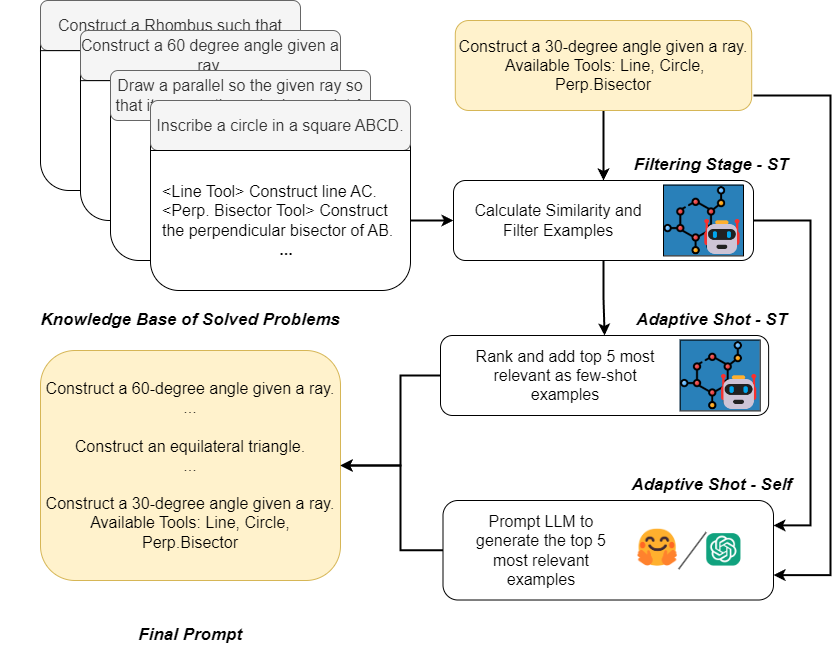}
    \caption{The adaptive few-shot mechanism. When facing the problem \textbf{\textit{Construct a 30-degree angle given a ray}}, we proceed to filter our knowledge base (top-right) and then either return the top five most similar results to build our prompt - Adaptive-Shot (ST) or prompt the LLM to filter them out by itself - Adaptive-Shot (Self).}
    \label{fig:adaptive}
\end{figure}

As seen in Figure \ref{fig:adaptive}, we begin by collecting a set of previously solved problems alongside their solutions, which call our knowledge base. This set can be acquired in multiple ways: In the case of our GSM8K / SVAMP / Geometry-Math experiments of Table \ref{tab:results_extra}, we use all problems belonging to the provided training split. In the case of Euclidea experiments of Table \ref{tab:results}, we begin with a small set of five handcrafted geometric challenges. Euclidea's problems vary in difficulty and are grouped into increasingly difficult level-packs. Since we are not provided with a training split, we add all problems belonging to previously seen packs to our knowledge base during the solution of a level. In this way, we continuously increase the size of our base as well as the availability of more diverse and complex techniques that our agents can utilize.\newline
The second stage involves the use of a Sentence Transformer to reduce the size of our knowledge base. For this, we compare the cosine similarity scores of our current problem and the problems in our knowledge base and keep entries with scores over $0.5$ or the top fifteen, whichever leads to less amount of examples.\newline
The third and final stage involves the use of the Sentence Transformer again, which chooses the top five most similar examples to build the final few-shot prompt. This procedure is referred to as Adaptive-Shot (ST) in our experiments. An alternative to this is to use the examples from the second step and prompt our solver-LLM to filter the top five most relevant examples by itself. We call this alternative Adaptive-Shot (Self). Note here that the second step of the procedure can be skipped during Adaptive-Shot (ST) since instead of filtering, we can directly return the top five results. However, it is necessary in the Adaptive-Shot (Self) method, where we need to respect the context size limits of the tested LLMs.

The following example presents three different few-shot prompts: A static-handcrafted version, an adaptive prompt compiled by a Sentence Transformer (Adaptive ST), and finally, an adaptive prompt filtered by the model itself (Adaptive Self). Here, our model is ChatGPT, and the problem at hand is:
\begin{dialog}[Problem]{gray}
Construct the circumcircle of a given triangle ABC.
\end{dialog}
\begin{dialog}[Available Tools]{gray}
[Line Tool, Circle Tool, Intersection Tool, Perpendicular Bisector Tool]
\end{dialog}
\begin{dialog}[Static Prompt \textcolor{red}{\xmark}]{red}
[...]\newline
Example 1:\newline
Description: Construct an equilateral triangle given one side. Let AB be the given side.\newline
Tool List: [Line Tool, Circle Tool, Intersection Tool, Perpendicular Bisector Tool]\newline
Solution:\newline
Circle Tool: Construct a circle with center A and radius AB.\newline
Circle Tool: Construct a circle with center B and radius AB, intersecting circle A at point C.\newline
Line Tool: Construct lines AC and BC. Triangle ABC is equilateral.
\newline

Example 2:\newline
Description: Construct a right angle at a point on a line. Given point A on a ray.\newline
Tool List: [Line Tool, Circle Tool, Intersection Tool, Perpendicular Bisector Tool]\newline
Solution:\newline
Circle Tool: Construct a circle with center A and an arbitrary radius, intersecting the ray at points B and C.\newline
Perpendicular Bisector Tool: Construct the perpendicular bisector of BC.
\newline

Example 3:\newline
Description: Bisect a given segment. Let AB be the given segment.\newline
Tool List: [Line Tool, Circle Tool, Intersection Tool]\newline
Solution:\newline
Circle Tool: Construct a circle with center A and radius AB.\newline
Circle Tool: Construct a circle with center B and radius AB, intersecting circle A at points C and D.\newline
Line Tool: Construct line CD, intersecting AB at point E. Point E is the midpoint of AB.
\newline

Example 4:\newline
Description: Construct a 45-degree angle given a right angle. Let angle XYZ be a right angle with vertex Y.\newline
Tool List: [Line Tool, Circle Tool, Intersection Tool, Perpendicular Bisector Tool]\newline
Solution:\newline
Circle Tool: Construct a circle with center Y and an arbitrary radius, intersecting XY at point A and YZ at point B.\newline
Perpendicular Bisector Tool: Construct the perpendicular bisector of AB, intersecting XY at point C and YZ at point D.\newline
Line Tool: Construct lines YC and YD. Angles XYC and YYD are 45 degrees.
\newline

Example 5:\newline
Description: Construct the perpendicular from a point to a line. Given point A and line l.\newline
Tool List: [Line Tool, Circle Tool, Intersection Tool, Perpendicular Bisector Tool]\newline
Solution:\newline
Circle Tool: Construct a circle with center A intersecting line l at points B and C.
\newline

\textbf{Problem Solution [ChatGPT]:}\newline
Circle Tool: Construct a circle at point A with radius equal to the distance between point A and any of the other two vertices, let's say point B.\newline
Circle Tool: Construct a circle at point B with radius equal to AB.\newline
Intersection Tool: Use the Intersection Tool to find the two intersection points between the two circles. These points are the circumcircle's endpoints.\newline
Line Tool: Connect the two intersection points to form the circumcircle of triangle ABC.\newline
\end{dialog}
\newpage
\thispagestyle{empty}
\begin{dialog}[Adaptive ST \checkmark]{orange}
[...]\newline
Example 1:\newline
Description: Inscribe a circle in the square.\newline
Tool List: [Line Tool, Circle Tool, Perpendicular Bisector Tool]\newline
Solution:\newline
Line Tool: Construct line AC.\newline
Perpendicular Bisector Tool: Construct the perpendicular bisector of AB, intersecting AB at E and AC at F.\newline
Circle Tool: Construct a circle with center E and radius EF.
\newline

Example 2:\newline
Description: Construct an equilateral triangle that is circumscribed about the circle O with radius R and contains the given point A on its circumference.\newline
Tool List: [Line Tool, Circle Tool, Perpendicular Bisector Tool, Perpendicular Tool]\newline
Solution:\newline
Line Tool: Construct line OA, intersecting the circle at B\newline
Perpendicular Tool: Construct the perpendicular to OA from A\newline
Circle Tool: Construct the circle with center B and radius OB, intersecting the circle at C and D, and line OA at E\newline
Line Tool: Construct line CE\newline
Line Tool: Construct line DE
\newline

Example 3:\newline
Description: Construct the midpoint of the segment that connects the midpoints of the diagonals of the quadrilateral.\newline
Tool List: [Line Tool, Circle Tool, Perpendicular Bisector Tool, Perpendicular Tool]\newline
Solution:\newline
Construct the perpendicular bisector of AB, intersecting AB at E\newline
Construct the perpendicular bisector of CD, intersecting CD at F\newline
Construct the perpendicular bisector of EF\newline
Construct line EF\newline

Example 4:\newline
Description: Inscribe an equilateral triangle in the circle using the given point as a vertex. The center of the circle is not given.\newline
Tool List: [Line Tool, Circle Tool, Perpendicular Bisector Tool, Perpendicular Tool]\newline
Solution:\newline
Circle Tool: Construct a circle with center A and an arbitrary radius, intersecting the original circle at B and C\newline
Circle Tool: Construct the circle with center B and radius AB, intersecting circle A at D and E\newline
Perpendicular Bisector Tool: Construct the perpendicular bisector of CD, intersecting the original circle at F\newline
Perpendicular Bisector Tool: Construct the perpendicular bisector of CE, intersecting the original circle at G\newline
Line Tool: Construct line FG
\newline

Example 5:\newline
Description: Circumscribe a square about the circle O. Two of its sides should be parallel to the given line.\newline
Tool List: [Line Tool, Circle Tool, Perpendicular Bisector Tool,  Perpendicular Tool]\newline
Solution:\newline
Perpendicular Tool: Construct the perpendicular to the line through O, intersecting the circle at A and B\newline
Perpendicular Tool: Construct the perpendicular to AB from A\newline
Perpendicular Tool: Construct the perpendicular to AB from B\newline
Circle Tool: Construct circle with center A and radius OA, intersecting the perpendicular to AB from A at C and D\newline
Perpendicular Tool: Construct the perpendicular to CD from C\newline
Perpendicular Tool: Construct the perpendicular to CD from D
\newline

\textbf{Problem Solution [ChatGPT]:}\newline
Perpendicular Bisector Tool: Construct the perpendicular bisector of side AB. Label the intersection with AB as D.\newline
Perpendicular Bisector Tool: Construct the perpendicular bisector of side BC. Label the intersection with BC as E.\newline
Perpendicular Bisector Tool: Construct the perpendicular bisector of side AC. Label the intersection with AC as F.\newline
Intersection Tool: Label the intersection of the perpendicular bisectors as O.\newline
Circle Tool: Create a circle with center O and radius OA.\newline
\end{dialog}
\newpage
\thispagestyle{empty}
\begin{dialog}[Adaptive Self \checkmark]{darkgreen}
[...]\newline
Example 1:\newline
Description: Construct a segment connecting the sides of the angle to get a triangle whose perpendicular bisectors are intersected in the point O.\newline
Tool List: [Line Tool, Circle Tool, Perpendicular Bisector Tool]\newline
Solution:\newline
Circle Tool: Construct the circle with center O and radius OA, intersecting the sides of the angle at B and C\newline
Line Tool: Construct line BC
\newline

Example 2:\newline
Description: Construct an equilateral triangle that is circumscribed about the circle O with radius R and contains the given point A on its circumference.\newline
Tool List: [Line Tool, Circle Tool, Perpendicular Bisector Tool, Perpendicular Tool]\newline
Solution:\newline
Line Tool: Construct line OA, intersecting the circle at B\newline
Perpendicular Tool: Construct the perpendicular to OA from A\newline
Circle Tool: Construct the circle with center B and radius OB, intersecting the circle at C and D, and line OA at E\newline
Line Tool: Construct line CE\newline
Line Tool: Construct line DE
\newline

Example 3:\newline
Description: Construct the midpoint of the segment that connects the midpoints of the diagonals of the quadrilateral.\newline
Tool List: [Line Tool, Circle Tool, Intersection Tool, Perpendicular Bisector Tool, Angle Bisector Tool, Perpendicular Tool]\newline
Solution:\newline
Perpendicular Bisector Tool: Construct the perpendicular bisector of AB, intersecting AB at E\newline
Perpendicular Bisector Tool: Construct the perpendicular bisector of CD, intersecting CD at F\newline
Perpendicular Bisector Tool:Construct the perpendicular bisector of EF
Line Tool: Construct line EF
\newline

Example 4:\newline
Description: Inscribe an equilateral triangle in the circle using the given point as a vertex. The center of the circle is not given.\newline
Tool List: [Line Tool, Circle Tool, Perpendicular Bisector Tool, Perpendicular Tool]\newline
Solution:\newline
Circle Tool: Construct a circle with center A and an arbitrary radius, intersecting the original circle at B and C\newline
Circle Tool: Construct the circle with center B and radius AB, intersecting circle A at D and E\newline
Perpendicular Bisector Tool: Construct the perpendicular bisector of CD, intersecting the original circle at F\newline
Perpendicular Bisector Tool: Construct the perpendicular bisector of CE, intersecting the original circle at G\newline
Line Tool: Construct line FG
\newline

Example 5:\newline
Description: Construct a segment connecting the sides of an angle to get a triangle whose orthocenter is in the given point O.\newline
Tool List: [Line Tool, Circle Tool, Perpendicular Bisector Tool,  Perpendicular Tool]\newline
Solution:\newline
Perpendicular Bisector Tool: Construct the perpendicular to one side R1 of the angle through O, intersecting the second side at A\newline
Perpendicular Bisector Tool: Construct the perpendicular to the second side R2 of the angle through O, intersecting the first side at B\newline
Line Tool: Construct line AB\newline

\textbf{Problem Solution [ChatGPT]:}\newline
Perpendicular Bisector Tool: Construct the perpendicular bisector of side AB. Label the intersection with AB as D.\newline
Perpendicular Bisector Tool: Construct the perpendicular bisector of side BC. Label the intersection with BC as E.\newline
Circle Tool: Create a circle with center D and radius AD.\newline
\end{dialog}

We observe that the static prompt leads to an incorrect solution, with the model misinterpreting the meaning of a circumcircle, and blindly performing step sequences similar to those presented in the prompt. 

In the case of Adaptive ST, the prompt consists of examples more closely related to the task at hand, leading to a correct solution. However, the model still suggests unnecessary steps, which can be attributed to style-mimicking tendencies introduced by instruction tuning. 

Finally, when the model is tasked to filter the examples by itself (Adaptive Self), we observe an overlap in the selected examples with the Sentence Transformer method. Here, the model suggests a correct sequence of steps, free from any unnecessary or repetitive tool usage.

\clearpage
\section{Target Variable Name Bias}
Below, we present a qualitative example of the effect of the target variable name on GPT4. The problem we present is: \textit{"Find a point that is equidistant from given points A and B."}. We present three illustrations, each signifying a different reasoning path to its solution, according to the name given to the target variable.

\begin{figure}[ht]
\centering
\includegraphics[width=0.45\textwidth]{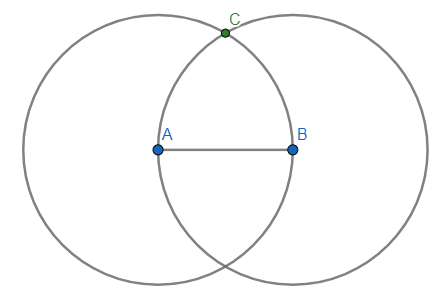}
\hfill
\includegraphics[width=0.45\textwidth]{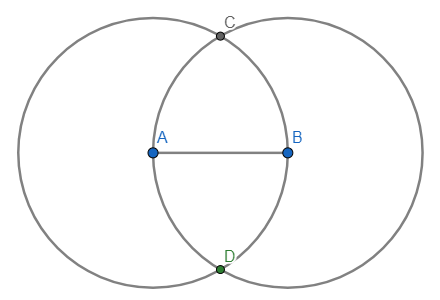}
\vspace{1em}
\includegraphics[width=0.45\textwidth]{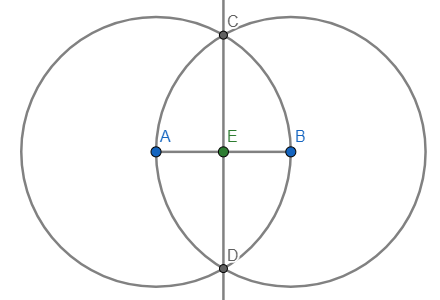}
\hfill
\includegraphics[width=0.45\textwidth]{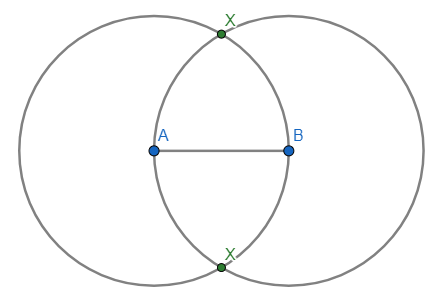}
\caption{Visualized GPT4 reasoning paths depending on the name of the target variable. Top left: $Target=C$, Top right: $Target=D$, Bottom left: $Target=E$ and Bottom right: $Target=X$}
\label{fig:fourSubplots}
\end{figure}

The model is prompted to solve the problem of finding a point that is equidistant from given points A and B. By altering the variable name of the target point, we observe different reasoning paths (Figure \ref{fig:fourSubplots}): Naming the target \textbf{C} leads to 3 steps with the point C being marked at the intersection of two circles. Naming the target \textbf{D} leads to the same solution, with D being identified as the second point of the intersection between the circles. Using \textbf{E} as the target urges the model to first involve the generation of the variables C and D in its solution (since they precede E in terms of alphabetical ordering). This leads to a different reasoning path: The model comes up with a 4-step solution, where E lies on line CD, the perpendicular bisector of line AB. However, an interesting case can be observed with the letter X, usually associated with a missing variable. Here, the model returns to the shorter 3-step reasoning and gives X multiple possible solutions. X is assigned to points rather than being explicitly constructed as the result of a tool.
\begin{table}[!h]
    \begin{mdframed}
\begin{dialog}[Problem]{gray}
Given two points, A and B, find a point \textbf{[Target]} so that its distance from A is equal to its distance from B.\newline
\end{dialog}
\begin{dialog}[Available Tools]{gray}
[Line Tool, Circle Tool, Intersection Tool, Perpendicular Bisector Tool]
\end{dialog}

\begin{dialog}[Target = C]{gray}
Circle Tool: Construct a circle with center A and radius AB. \newline
Circle Tool: Construct another circle with center B and radius AB. \newline
Intersect Tool: Identify the points where the two circles intersect. These intersection points are equidistant from A and B. Let's call one of these intersection points C.\newline
The point C, found at the intersection of the two circles, satisfies the condition that its distance from A is equal to its distance from B.\newline
\end{dialog}

\begin{dialog}[Target = D]{gray}
Circle Tool: Construct a circle with center A and radius AB. \newline
Circle Tool: Construct a circle with center B and radius AB. \newline
Intersect Tool: Find the intersection points of the two circles, let's call them C and D. These points are equidistant from A and B. \newline
By following these steps, point D will be at an equal distance from both points A and B. \newline
\end{dialog}

\begin{dialog}[Target = E]{gray}
Circle Tool: Use the circle tool to draw a circle with center A and radius AB. \newline
Circle Tool: Use the circle tool again to draw a circle with center B and radius AB. \newline
Intersect Tool: The two circles will intersect at two points, which we can call point C and point D.\newline
Line Tool: Draw the line CD.\newline
Choose any point on line CD (other than C or D) and label it point E. \newline
\end{dialog}

\begin{dialog}[Target = X]{gray}
Circle Tool: Construct a circle with center A and radius AB. \newline
Circle Tool: Construct a circle with center B and radius AB. \newline
Intersection Tool: The intersection points of the two circles represent the points that are equidistant from A and B. These intersection points are the solutions for point X. \newline
In this case, \textbf{there will be two solutions for X}, each lying on the circumference of the circles where they intersect. \newline
\end{dialog}
\end{mdframed}
    \caption{The completions of the model for the above problem, in geometric tool language.}
    \label{example: gpt4Bias}
\end{table}

\clearpage
\section{Baselines}
We introduce two baselines to analyze the complexity of the geometric construction problem. In the first baseline, which we will call "Longest Common Solution" (LCS), we identify the top five longest common sequences of steps between ground truth solutions. Then, for each given problem, we uniformly sample from these sequences, adjusting the variables of each step and the tool usage to the current task. If a sampled step does not apply to the current state of the problem, the sampled sequence is discarded, and a new one is sampled instead. This baseline corresponds to the success rate of an agent who memorized a set of fixed solution steps and applied them to each given problem.

Additionally, we propose a second baseline named 'N-Gram Rollouts' (N-Gram), which begins by creating a database of uni-, bi-, and tri-grams derived from the tools used in ground-truth answers. For each problem, our method involves a two-phase iterative process. Initially, we choose either a single tool ($n=1$) or a sequence of tools ($n>1$) from our database. Following this, we select the geometric variables upon which these tools will be applied. To facilitate this, we maintain a memory initially populated with variables given in the problem statement. It's important to note that tool application varies in complexity, with some tools requiring a single variable (e.g., constructing a ray from point A) and others necessitating two (e.g., drawing a line between points A and B). For each tool or sequence of tools selected, we sample the required number of variables from memory, with recent variables weighted more heavily than older ones, following an exponential decay schema. Any new variables a tool generates (such as a new point) are added to this memory. This process is repeated until a predefined number of steps is reached.

\begin{table}[ht]
\centering
\begin{tabular}{lcc}
\toprule
\textbf{Method} & \textbf{Correct tool sequence} & \textbf{Fully correct solution} \\ \midrule
LCS             & 0.17                                     & 0.02 \\ 
1-Gram          & 0.14                                     & 0.11 \\ 
2-Gram          & 0.12                                     & 0.09 \\ 
3-Gram          & 0.07                                     & 0.06  \\
\bottomrule
\end{tabular}
\caption{Performance of Baselines. Results refer to pass@1 estimated after 1000 trials.}
\label{tab:baselines}
\end{table}

According to Table \ref{tab:baselines}, finding an entirely correct solution randomly is almost  0.1\% probable. Note here that finding the correct tool sequence score is significantly higher. However, this is expected since many problems share common starting steps (e.g., drawing a line or a circle).
\clearpage

\section{Visual Relations Prompt}
\label{sec:vrp}
\begin{figure}[!h]
    \centering
      \includegraphics[width=0.87\textwidth]{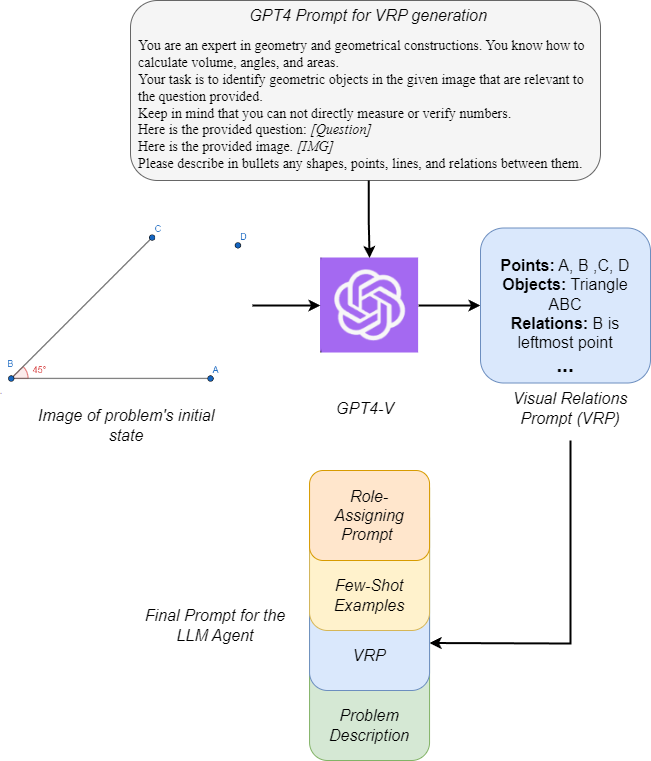}
    \caption{VRP extraction using GPT4-V}
    \label{fig:vrp}
\end{figure}

In Figure \ref{fig:vrp}, we provide a visualization of our proposed visual relations prompt:\newline

We begin by employing an auxiliary prompt that contains the problem question and an image depicting the initial state of the problem. This is presented to GPT4-V, where a list of \textbf{\textit{Points}}, \textbf{\textit{Lines}}, \textbf{\textit{Objects}}, and their \textbf{\textit{Relations}} is returned in the form of bullet points. This information, which we refer to as VRP, is then added to the overall agent prompt between the few-shot examples and the current problem and available tool description.

\clearpage
Here are two examples presenting the benefits of VRP inclusion.

\textbf{Example 1:}\newline
\begin{mdframed}
\begin{dialog}[Problem]{gray}
Given the rectangle, ABCD, and the point E located over the side AB at a random position, construct a line through E that cuts the rectangle into two parts of equal area.
\end{dialog}
\begin{dialog}[Available Tools]{gray}
[Line Tool, Circle Tool, Perpendicular Bisector Tool, Angle Bisector Tool, Intersect Tool]\newline
\end{dialog}
\begin{dialog}[MetaMath-Mistral]{gray}
Line Tool: Construct line EF, parallel to AB.\newline
Perpendicular Bisector Tool: Construct the perpendicular bisector of line EF.\newline
Intersect Tool: Mark the intersection of the perpendicular bisector and line AB as point G.\newline
Line Tool: Construct line EG.\textcolor{red}{\textcolor{red}{\xmark}} 
\end{dialog}
\begin{dialog}[ChatGPT]{gray}
Perpendicular Bisector: Construct the perpendicular bisector of AB. This bisector will cut the rectangle into two equal parts.\newline
Intersect Tool: Extend this perpendicular bisector line to intersect side CD. Label this intersection point as F. \newline
Draw a line through point E and point F. This line will cut the rectangle into two parts of equal area. \textcolor{red}{\textcolor{red}{\xmark}} 
\end{dialog}
\begin{dialog}[GPT4]{gray}
Line Tool: Construct Line AE.\newline
Perpendicular Bisector: Construct the perpendicular bisector of CD.\newline
Intersect Tool: Mark the intersection of the perpendicular bisector with CD and mark it as H.
This point H is the midpoint of CD.\newline
Line Tool: Draw a line from point E to point H. \textcolor{red}{\textcolor{red}{\xmark}} 
\end{dialog}
\captionsetup[figure]{labelformat=customdialogue}
\end{mdframed}
\textbf{Example 2:}\newline
\begin{mdframed}
\begin{dialog}[Image]{gray}
\centering
\includegraphics{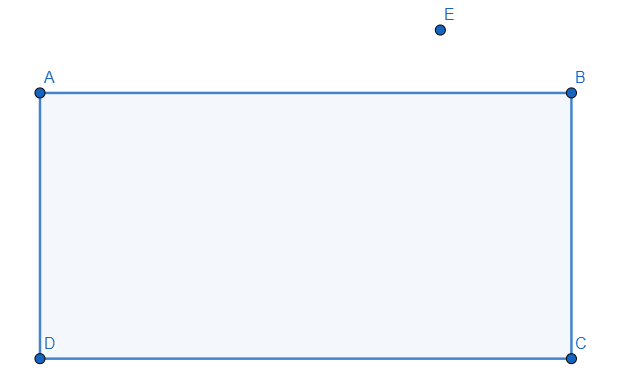}
\end{dialog}
\begin{dialog}[Problem]{gray}
Given the rectangle, ABCD, and point E located over the side AB at a random position, construct a line through E that cuts the rectangle into two parts of equal area.
\end{dialog}
\begin{dialog}[Available Tools]{gray}
[Line Tool, Circle Tool, Perpendicular Bisector Tool, Angle Bisector Tool, Intersect Tool]\newline
\end{dialog}
\begin{dialog}[Visual Relations Prompt]{gray}
Here is the full description of all the relations between geometric objects:\newline
\textbf{Shapes:}\newline
Rectangle: ABCD\newline
\textbf{Points}:\newline
A: Top-left corner of the rectangle.\newline
B: Top-right corner of the rectangle.\newline
C: Bottom-right corner of the rectangle.\newline
D: Bottom-left corner of the rectangle.\newline
E: Point outside of the rectangle.\newline
\textbf{Lines}:\newline
AB: Top horizontal line of the rectangle.\newline
BC: Right vertical line of the rectangle.\newline
CD: Bottom horizontal line of the rectangle.\newline
DA: Left vertical line of the rectangle.\newline
\textbf{Relations}:\newline
AB is parallel to CD.\newline
AD is parallel to BC.\newline
The length of AB is equal to the length of CD.\newline
The length of AD is equal to the length of BC.\newline
E is an isolated point, not connected to any lines of the rectangle ABCD.\newline
\end{dialog}
\begin{dialog}[MetaMath-Mistral]{gray}
Perpendicular Bisector Tool: Construct the perpendicular bisector of the line AB.\newline
Perpendicular Bisector Tool: Construct the perpendicular bisector of the line CD.\newline
Perpendicular Bisector Tool: Construct the perpendicular bisector of the line BC.\newline
Perpendicular Bisector Tool: Construct the perpendicular bisector of the line DA.\newline
Intersect Tool: Find the intersection of the perpendicular bisectors of AB and CD and name it F.\newline
Intersect Tool: Find the intersection of the perpendicular bisectors of BC and DA and name it G.\newline
Line Tool: Draw a line from E to F.\textcolor{darkgreen}{\textcolor{darkgreen}{\cmark}}
\end{dialog}
\begin{dialog}[ChatGPT]{gray}
Line Tool: Draw a diagonal of the rectangle AC. \newline
Perpendicular Bisector Tool: Construct the perpendicular bisector of AC. \newline
Intersect Tool: Label the intersection point of the perpendicular bisector and AC as M. \newline
Line Tool: Draw a line segment from point E to point M. This line, EM, will cut the rectangle ABCD into two parts of equal area since it passes through the midpoint of the diagonal AC.\textcolor{darkgreen}{\textcolor{darkgreen}{\cmark}}
\end{dialog}
\begin{dialog}[GPT4]{gray}
Line Tool: Draw a diagonal of the rectangle AC. \newline
Perpendicular Bisector Tool: Construct the perpendicular bisector of AC. \newline
Intersect Tool: Label the intersection point of the perpendicular bisector and AC as M. \newline
Line Tool: Draw a line segment from point E to point M. This line, EM, will cut the rectangle ABCD into two parts of equal area since it passes through the midpoint of the diagonal AC.\textcolor{darkgreen}{\textcolor{darkgreen}{\cmark}}
\end{dialog}
\end{mdframed}
When tasked to tackle the same geometrical problem, our three agents (MetaMath-Mistral / ChatGPT / GPT4)  resort to a "convenient" placement of point E outside rectangle ABCD. All of them try to utilize the fact that the perpendicular bisector of any edge of a rectangle cuts the rectangle in two equal areas. Then a series of erroneous steps are suggested, trying to connect point E with this line. 
After the introduction of the VRP, we observe that all agents manage to correct their responses. The placement of geometric objects in the prompt seems to assist them in understanding the nature of the problem better. Agents are no longer resorting to steps that simply suggest splitting the rectangle area into two equal parts, ignoring point E, but rather respect the special spatial circumstances the problem presents. Note here that the initial solutions are valid if the random position of point E happens to align with the midpoint of any side. The initial tendency to produce such solutions can be attributed to the fact that the splitting of a rectangle must have been observed more frequently in their training data through the use of diagonals or lines connecting midpoints.

\section{Multi-Agent Dialogue Examples}
\label{sec:mad}
Illustration of our proposed $SV_{NL}-SV_{GT}$ multi-agent setup. We present the agent dialogue leading to a successful solution. The use of a single ChatGPT instance is not sufficient to solve this particular problem. However, the combined power of multiple agents can. 

\begin{center}
According to our proposed method, we employ 4 agents: 
\begin{enumerate}
\centering
\item Natural Language Solver \hspace{8pt}\inlineimg{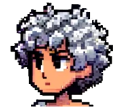}(\textcolor{blue}{NL Solver}) 

\item Natural Language Validator \inlineimg{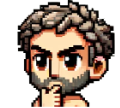}(\textcolor{darkgreen}{NL Validator}) 

\item Geometric Tools Solver \hspace{10pt}\inlineimg{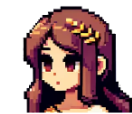}(\textcolor{red}{GT Solver})

\item Geometric Tools Validator \hspace{5pt} \inlineimg{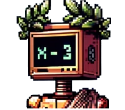}(\textcolor{orange}{GT Validator})
\end{enumerate} 
\end{center}

The four different agents interact with each other from top to bottom and from left to right.
\begin{mdframed}
\begin{dialog}[Problem]{gray}
Given the square ABCD, inscribe a circle in the square.
\end{dialog}
\begin{dialog}[Available Tools]{gray}
[Line Tool, Circle Tool, Perpendicular Bisector Tool, Intersect Tool]
\end{dialog}
\noindent
\begin{minipage}[c]{\textwidth} 
\begin{dialog}[NL Solver \inlineimg{images/nl_solver.png}]{blue}
[...] To inscribe a circle in the square ABCD means to draw a circle such that all four sides of the square are tangent to the circle. The circle should touch each side of the square at exactly one point. I suggest the following steps:
\newline
\step 1: Draw the square ABCD. \newline
\step 2: Draw the diagonal AC of the square. The diagonal AC will pass through the center of the circle.\newline
\step 3: Construct the perpendicular bisector of the diagonal AC. This line will intersect the diagonal at the center of the circle. \newline
\step 4: Draw a line segment from the center of the circle to one of the vertices of the square (e.g., A or C). This line will be the radius of the circle. \newline
\step 5: Use the compass to measure the length of the line segment drawn in the previous step. \newline
\step 6: Set the compass to the measured length and draw a circle with the center at the intersection of the diagonal AC and the perpendicular bisector.\newline\newline
\end{dialog}
\end{minipage}

\vspace{2mm} 

\noindent 
\begin{minipage}[c]{\textwidth}
\begin{dialog}[NL Validator \inlineimg{images/nl_validator.png}]{darkgreen}
According to the task and the provided rationale, I am confident regarding the validity of the following steps: \newline
\step 1: Correct. \newline
\step 2: Correct. \newline
\step 3: Correct. \newline
Please try to correct the remaining steps. \newline\newline
\end{dialog}
\end{minipage}

\vspace{2mm} 
\noindent 
\begin{minipage}[c]{\textwidth} 
\begin{dialog}[NL Solver\inlineimg{images/nl_solver.png}]{blue}
[...]\step 4: Draw a line segment from the center of the circle to one of the points where the circle is tangent to the square. This line segment will be the radius of the circle. \newline

Following these corrected steps, we will successfully inscribe a circle in the square ABCD.\newline\newline
\end{dialog}
\end{minipage}

\vspace{2mm} 
\noindent
\begin{minipage}[c]{\textwidth} 
\begin{dialog}[GT Solver\inlineimg{images/gt_solver.png}]{red}
Given the suggested expert rationale and the available tools I suggest the following steps:\newline
\lt 1: Draw a line segment from point A to point B.\newline
\lt 2: Draw a line segment from point A to point C.\newline
\pbt 3: Draw the perpendicular bisector of AB.\newline
\inrt 4: Mark the intersection point of the bisector and AB as E.\newline
\inrt 5: Mark the intersection of the perpendicular bisector and the diagonal AC as O.\newline
\ct 6: Create a circle with center O and radius AO.\newline\newline
\end{dialog}
\end{minipage}

\vspace{2mm} 
\noindent 
\begin{minipage}[c]{\textwidth}
\begin{dialog}[GT Validator\inlineimg{images/gt_validator.png}]{orange}
According to the steps given, the available tools and the suggested rationale, I am confident of the validity of the following steps: \newline
\lt 1: Correct.\newline
\lt 2: Correct.\newline
\pbt 3: Correct.\newline
\inrt 4: Correct.\newline
\inrt 5: Correct.\newline
Please try to correct the solution according to my suggestions.\newline\newline
\end{dialog}
\end{minipage}

\vspace{2mm} 
\noindent
\begin{minipage}[c]{.45\textwidth}
\begin{dialog}[GT Solver\inlineimg{images/gt_solver.png}]{red}
[...]
\ct 6: Create a circle with center O and radius OE. \textcolor{darkgreen}{\cmark}\newline\newline
\end{dialog}
\end{minipage}
\vfill
\hfill \footnotesize \textcolor{gray}{ChatGPT Simulaca}
\label{examples:correct_simulacra_ChatGPT}
\end{mdframed}
\clearpage

Below, we present another example where we showcase the benefits of incorporating variable renaming in multi-agent scenarios.
Simulacra are not immune to the variable name bias: In the example below, the variables \textbf{A}, \textbf{B}, and \textbf{C} are used. \textbf{A} and \textbf{B} mark the two initial points of the problem, while \textbf{C} stands as the target to be found. Initially, the \textcolor{blue}{NL Solver} comes up with a promising 5-step rationale. However, the {\textcolor{darkgreen}{NL Validator}} corrects the proposed final step as a mistake. Since \textbf{A},\textbf{B}, and \textbf{C} are consecutive letters in the alphabet, and \textbf{C} has already appeared in the fourth step, the validator marks the final step as redundant.
The new variable \textbf{D} generated at the fifth step is after \textbf{C} (in terms of alphabetical ordering), and thus the problem must have been already solved.  This error is propagated to the GT agents, leading to an eventually wrong final response. 

\begin{mdframed}
\begin{dialog}[Problem]{gray}
Let $|AB|=1$. Construct a point C on the line AB such that the length of AC is equal to $\sqrt{2}$.Do not use arbitrary numbers but only existing lengths and sizes in your solution\newline
\end{dialog}
\begin{dialog}[Available Tools]{gray}
[Line Tool, Circle Tool, Perpendicular Bisector Tool, Perpendicular Tool, Angle Bisector Tool, Intersect Tool]\newline
\end{dialog}
\noindent
\begin{minipage}[c]{\textwidth} 
\begin{dialog}[NL Solver\inlineimg{images/nl_solver.png}]{blue}
[...] To create a line segment whose length is the square root of 2, we should utilize an existing segment of length 1, here AB. This requires forming a right-angled triangle where one side measures 1, and determining the hypotenuse's length which will be the square root of 2. I suggest the following steps:\newline
\step 1: Draw a straight line segment AB of any length. This will serve as our unit length. \newline
\step 2: At point B, construct a perpendicular line to AB. \newline
\step 3: From point B, draw a line segment BC on the perpendicular line with the same length as AB. \newline
\step 4: Connect points A and C to form a right triangle ABC. The length of AC, the hypotenuse, is the square root of 2. \newline
\step 5: Finally, mark the point D on the line AB such that AD = AC This will give us the line segment AD of length square root of 2. \textcolor{darkgreen}{\cmark}\newline\newline
\end{dialog}
\end{minipage}

\vspace{2mm} 

\noindent 
\begin{minipage}[c]{\textwidth}
\begin{dialog}[NL Validator\inlineimg{images/nl_validator.png}]{darkgreen}
According to the task and the provided rationale, I am confident regarding the validity of the following steps: \newline
\step 1: Correct. \newline
\step 2: Correct. \newline
\step 3: Correct. \newline
\step 4: Correct. \newline
Please try to correct the solution according to my suggestions.\newline\newline
\end{dialog}
\end{minipage}

\vspace{2mm} 
\noindent 
\begin{minipage}[c]{\textwidth} 
\begin{dialog}[NL Solver\inlineimg{images/nl_solver.png}]{blue}
I suggest we should re-write the steps so that all suggested changes are incorporated.\newline
[...]
\step 5: Point C is the required point on the line segment extended from AB such that the length AC equals the square root of 2.\textcolor{red}{\xmark}\newline\newline
\end{dialog}
\end{minipage}

\vspace{2mm} 
\noindent
\begin{minipage}[c]{\textwidth} 
\begin{dialog}[GT Solver\inlineimg{images/gt_solver.png}]{red}
Given the suggested expert rationale and the available tools I suggest the following steps:\newline
\ct 1: Use the circle tool to draw a circle with center A and radius AB.\newline
\pt 2: Draw a perpendicular line to AB at point B using the Perpendicular Tool.\newline
\ct 3: Draw a circle with center B and radius BA.\newline
\inrt 4: Use the Intersection Tool to find point C, which is the intersection of the circle centered at B and the line perpendicular to AB.\newline
\lt 5: Draw a line from A to C using the Line Tool. \textcolor{red}{\xmark}\newline
\end{dialog}
\end{minipage}

\vspace{2mm} 
\noindent 
\begin{minipage}[c]{.45\textwidth}
\begin{dialog}[GT Validator\inlineimg{images/gt_validator.png}]{orange}
According to the steps given, the available tools and the suggested rationale, I am confident of the validity of the following steps: \newline
\ct 1: Correct. \newline
\pt 2: Correct.\newline
\ct 3: Correct.\newline
\inrt 4: Correct.\newline
\lt 5: Correct.\newline

Great job!
\end{dialog}
\end{minipage}
\hfill 
\begin{minipage}[c]{.45\textwidth}
\begin{dialog}[GT Solver\inlineimg{images/gt_solver.png}]{red}
Thank you!\newline
\end{dialog}
\end{minipage}
\vfill
\hfill \footnotesize \textcolor{gray}{GPT4 Simulacra}
\label{examples:symbol_issue}
\end{mdframed}

Now, let us rename the target variable and present the problem to our multi-agent setup again. Instead of the consecutive letter \textbf{C}, we swap the final target with a far-away (in terms of alphabetical ordering) as well as a common mathematical symbol \textbf{X}. We observe that the validator now approves the suggested rationale, leading to a successful generation.

\noindent
\begin{mdframed}
\begin{dialog}[Target Variable Renaming]{gray}
Let $|AB|=1$. Construct a point \textcolor{red}{X} on the line AB such that the length of \textcolor{red}{AX} is equal to $\sqrt{2}$. Do not use arbitrary numbers but only existing lengths and sizes in your solution.\newline
\end{dialog}
\noindent
\begin{minipage}[c]{\textwidth} 
\begin{dialog}[NL Solver\inlineimg{images/nl_solver.png}]{blue}
[...] Based on the problem description, we are tasked with constructing a point on a given line segment such that the length of the segment from the initial point to the new point equals the square root of 2 without using explicit numerical values. Here's how we can approach this:\newline
\step 1: Draw a straight line segment AB of any length (this will be our unit length as we are not to use numbers).\newline
\step 2: Construct a circle with center at A and radius AB.\newline
\step 3: Draw a line perpendicular to AB at point B. This line intersects the circle A at a new point D.\newline
\step 4: Draw a line segment AD. The length of AD is the square root of 2 (according to Pythagoras' theorem).\newline
\step 5: Take point X such that AX = AD. Now AX is equal to the square root of 2.\textcolor{darkgreen}{\cmark}\newline
\end{dialog}
\end{minipage}

\vspace{2mm} 

\noindent 
\begin{minipage}[c]{\textwidth}
\begin{dialog}[NL Validator\inlineimg{images/nl_validator.png}]{darkgreen}
According to the task and the provided rationale, I am confident regarding the validity of the following steps: \newline
\step 1: Correct. \newline
\step 2: Correct. \newline
\step 3: Correct. \newline
\step 4: Correct. \newline
\step 5: Correct. \newline
Your solution is valid!
\end{dialog}
\end{minipage}

\vspace{2mm}
\noindent 
\begin{minipage}[c]{\textwidth} 
\begin{dialog}[NL Solver\inlineimg{images/nl_solver.png}]{blue}
Thank you for your help!\newline
\end{dialog}
\end{minipage}

\vspace{2mm} 

\noindent
\begin{minipage}[c]{\textwidth} 
\begin{dialog}[GT Solver\inlineimg{images/gt_solver.png}]{red}
Given the suggested expert rationale and the available tools I suggest the following steps:\newline
\lt 1: Use the line tool to draw the line AB.\newline
\ct 2: Draw a circle with center B and radius BA.\newline
\pt 3: Draw a perpendicular line at point B.\newline
\inrt 4: Use the Intersection Tool to find point D, which is the intersection of the circle centered at B and the line perpendicular to AB.\newline
\lt 5: Draw the line AD.\newline
\lt 6: Draw a line from A to X. \textcolor{red}{\xmark}
\end{dialog}
\end{minipage}

\vspace{2mm} 
\noindent 
\begin{minipage}[c]{.45\textwidth}
\begin{dialog}[GT Validator\inlineimg{images/gt_validator.png}]{orange}
According to the steps given, the available tools and the suggested rationale, I am confident of the validity of the following steps: \newline
\lt 1: Correct. \newline
\ct 2: Correct.\newline
\pt 3: Correct.\newline
\inrt 4: Correct.\newline
\lt 5: Correct.\newline
Please try to correct the solution according to my suggestions.\newline\newline
\end{dialog}
\end{minipage}
\hfill 
\begin{minipage}[c]{.45\textwidth}
\begin{dialog}[GT Solver \inlineimg{images/gt_solver.png}]{red}
[...] \ct 6: Draw a circle with center A and radius AD. \newline
\inrt 7: Mark the point of intersection of the circle with the line AB as X. X is the desired point. \textcolor{darkgreen}{\cmark}\newline\newline
\end{dialog}
\end{minipage}
\vfill
\hfill \footnotesize \textcolor{gray}{GPT4 Simulacra}
\label{examples:variable_renaming}
\end{mdframed}
\clearpage
\end{document}